# Image Interpolation Using Kriging Technique for Spatial Data

Firas Ajil Jassim , Fawzi Hasan Altaany

**Abstract** —*Image interpolation has been used spaciously by customary interpolation techniques. Recently, Kriging technique has been widely implemented in simulation area and geostatistics for prediction. In this article, Kriging technique was used instead of the classical interpolation methods to predict the unknown points in the digital image array. The efficiency of the proposed technique was proven using the PSNR and compared with the traditional interpolation techniques. The results showed that Kriging technique is almost accurate as cubic interpolation and in some images Kriging has higher accuracy. A miscellaneous test images have been used to consolidate the proposed technique.[1]*

## I. INTRODUCTION

Image interpolation is a significant operation in image processing which can be used to resample the image either to decrease or increase the resolution [10]. The most main problems that may face the researcher in image interpolation are the quality of the resulted image and the computational complexity required to compute the interpolated image [3]. Several commonly used interpolation algorithms have been suggested, such as nearest neighbor interpolation, linear interpolation and cubic spline interpolation [8], [11]. Earlier researchers used to use cubic spline interpolation as a very sophisticated method to interpolate an image. Mathematically, spline refers to a piecewise function consisting of polynomial pieces. It is commonly used to describe curves and surfaces in computer-aided design and related fields. For signal and image processing, spline interpolation has been considered as a useful tool and been used intensively in various tasks [12], [7]. A spline is a polynomial between n-points where its order is $O(n-1)$. Actually, increasing the order of the polynomial does not necessarily increase the accuracy of the resulted image [2].

Additionally, image interpolation can be useful when the aim is to decrease or increase the file size but this must not affect the quality of the resulted image too much. Also, the reconstructed image differences must not be noticeable by the human eye [15].

## II. KRIGING BACKGROUND

Kriging is an alternative to many other point interpolation techniques. Kriging is an interpolation method that can produce predictions of unobserved values from observations of its value at nearby locations. Kriging confer weights for each point according to its distance from the unknown value. Actually, these predictions treated as weighted linear combinations of the known values. Kriging method is more accurate whenever the unobserved value is closer to the observed values [13]. The prediction obtained by Kriging method is more accurate than polynomial interpolation [14]. Note that not much study related to Kriging in image processing or compression has been reported, which the most recent can be found in [1], [4] and [6].

The basic form of the Kriging estimator is that:

$$P^* = \sum_{i=1}^{n} \lambda_i P_i \qquad (1)$$

Where $P_1, P_2, ..., P_n$ are the observed points and $P^*$ is the point where we want to predict its value, also we may note that:

$$\sum_{i=1}^{n} \lambda_i = 1 \qquad (2)$$

Actually, the goal is to determine weights, $\lambda_i$'s that minimize the variance of the estimator, [13], [9], under the unbiasedness constraint:

$$E\{P^* - P\} = 0 \qquad (3)$$

There are several Kriging methods, differ in their treatments of the weighted components $\lambda_i$'s. Here, in the proposed technique, ordinary kriging will be used due to the fact that it is the most common kriging type used in simulation for spatial data and it is considered to be best because it minimizes the variance of the estimation error [13].

[1] Firas Ajil Jassim is with Irbid National University, Irbid, Jordan (e-mail: firasajil@yahoo.com).
Fawzi Hasan Altaany with Irbid National University, Irbid, Jordan (e-mail: fawzitaani@yahoo.com).

### III. Grid Methods

As a traditional pattern, image interpolation was implemented by accreditation an 8×8 block from the digital image array. In order to implement the interpolation as a prediction criterion, a curtailment procedure may be applied to curtail some values from the original 8×8 block for the purpose of predicting them later. In this paper, three different scenarios have been suggested to predict the curtailed values which are: horizontal grid, vertical grid and 8-points grid. According to figure I, the black squares represent the original pixels from the original 8×8 block while the white squares are the missing (curtailed) pixels that are to be interpolated. Obviously, for both the horizontal and vertical grids, the unobserved pixels for each 8×8 are 32 which are the same as the observed pixels. Therefore, 50% are observed and 50% are not. Therefore, the interpolation is made by 50% of the pixels for both the horizontal and vertical grids. In the 8-points grid, 8 original values out of 64 values remain as they are in the original grid while 56 values omitted. The prediction was made by (8/64) = 12.5% which is very low percentage. Logically, whenever the number of unknown points increased and the value of the error metric is better, this means that the prediction method is more accurate.

Horizontal grid

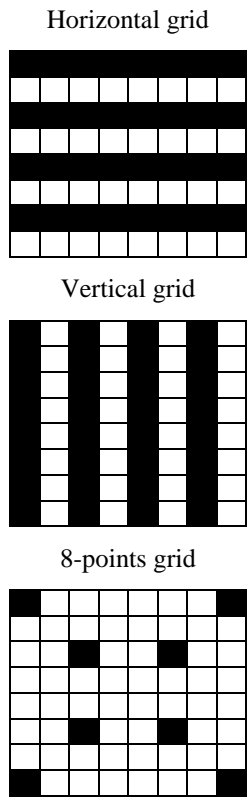

Vertical grid

8-points grid

**Fig 1. Horizontal, vertical and 8-points grids**

Clearly, decreasing the number of observed points (black) in the gridding method will make the interpolation more difficult to evaluate. Therefore, 8-points gridding method is the most convenient method that will make it hard to interpolate unobserved points precisely. Hence, the 8-points grid is the harder grid interpolation method that could be used to measure the exact efficiency between two methods, i.e. traditional interpolation and Kriging method for interpolation.

### IV. EXPERIMENTAL RESULTS

As an application of Kriging in image interpolation, the spatial domain of an image can be very suitable to implement and accommodate Kriging method. Firstly, the 8×8 block is used as a standard block size to interpolate the unknown values either by horizontal, vertical or 8-points gridding method. The most three popular interpolation methods: nearest-neighbor, linear and cubic spline was evaluated and compared with Kriging method using the PSNR (peak signal-to-noise ratio) [5]. In this article, fifteen test images (512×512) have been tested and compared between traditional interpolation and Kriging method, figure V. First of all, instead of applying the proposed method to the overall image array, we can test it for a single 8×8 block. An illustrative example showing the pixel values from Lena test image are presented. A random 8×8 block has been chosen to show the prediction results obtained from both cubic spline and Kriging methods, Fig II, III and IV.

| | | | | | | | |
|---|---|---|---|---|---|---|---|
| 211 | 208 | 209 | 211 | 211 | 208 | 208 | 211 |
| 210 | 209 | 209 | 211 | 210 | 208 | 208 | 209 |
| 210 | 211 | 211 | 211 | 210 | 209 | 208 | 208 |
| 209 | 211 | 212 | 210 | 208 | 209 | 208 | 207 |
| 207 | 209 | 210 | 208 | 207 | 208 | 208 | 206 |
| 207 | 209 | 210 | 208 | 207 | 208 | 209 | 208 |
| 208 | 209 | 210 | 209 | 208 | 207 | 209 | 210 |
| 208 | 208 | 208 | 208 | 206 | 205 | 207 | 209 |

**Fig II. Original 8×8 block**

| | | | | | | | |
|---|---|---|---|---|---|---|---|
| 211 | 211 | 211 | 210 | 210 | 210 | 210 | 211 |
| 210 | 211 | 210 | 209 | 209 | 209 | 209 | 210 |
| 210 | 210 | 211 | 210 | 209 | 209 | 209 | 209 |
| 209 | 210 | 210 | 210 | 209 | 208 | 208 | 209 |
| 209 | 210 | 210 | 209 | 208 | 208 | 208 | 209 |
| 208 | 209 | 210 | 209 | 208 | 208 | 208 | 209 |
| 208 | 209 | 208 | 208 | 207 | 207 | 208 | 209 |
| 208 | 208 | 208 | 208 | 208 | 208 | 208 | 209 |

**Fig III. Predicted 8×8 block using Cubic spline**

| | | | | | | | |
|---|---|---|---|---|---|---|---|
| 210 | 211 | 210 | 210 | 210 | 210 | 210 | 210 |
| 210 | 210 | 210 | 210 | 210 | 209 | 210 | 210 |
| 210 | 210 | 211 | 210 | 209 | 208 | 209 | 209 |
| 210 | 210 | 210 | 209 | 209 | 208 | 208 | 209 |
| 209 | 210 | 210 | 209 | 208 | 208 | 208 | 208 |
| 209 | 209 | 210 | 209 | 208 | 208 | 208 | 208 |
| 208 | 209 | 209 | 208 | 208 | 208 | 208 | 208 |
| 208 | 208 | 208 | 208 | 208 | 208 | 208 | 209 |

**Fig IV. Predicted 8×8 block using Kriging method**

Mathematically speaking, the error between the two blocks can be measured using PSNR. The PSNR for Cubic spline was (44.8113) while for Kriging it was (45.0461).

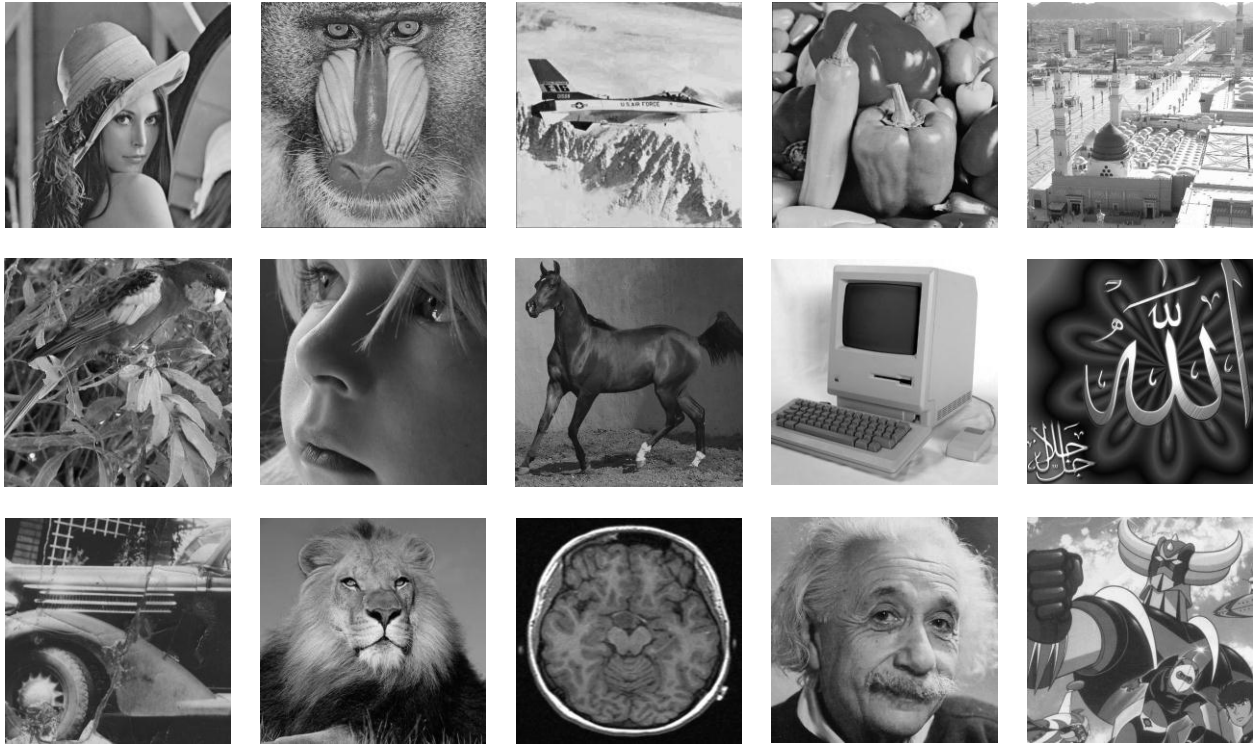

**Fig. V. 512×512 test images: Lena, baboon, f16, peppers, mosque, bird, baby, horse, mac, Allah, oldcar, lion, MRI, Einstein and Grendizer**

Therefore, since the value of PSNR of Kriging is higher than Cubic spline, this implies that Kriging produce better results than Cubic spline. The PSNR value for the fifteen test images have been computed for the three gridding types (horizontal, vertical and 8-points) and showed in table I, II and III.

As seen from tables I and II, the values of PSNR is higher for the traditional interpolation methods than Kriging method. The convergence between horizontal and vertical grid pixels make it easy to interpolate for the traditional interpolation method while makes it difficult for Kriging method. On the other hand, in table III the value of PSNR for Kriging is higher compared with the traditional interpolation.

**TABLE I**
**HORIZONTAL GRID**

|  | Nearest | Linear | Cubic | Kriging |
|---|---|---|---|---|
| Lena | 34.40 | 41.69 | 42.63 | 41.01 |
| Baboon | 22.53 | 24.32 | 24.60 | 24.25 |
| F16 | 31.68 | 38.24 | 39.08 | 38.09 |
| Peppers | 30.92 | 36.25 | 34.43 | 36.01 |
| Mosque | 23.56 | 25.31 | 25.19 | 25.25 |
| Bird | 29.26 | 32.84 | 33.08 | 32.81 |
| Baby | 36.24 | 38.82 | 38.90 | 38.38 |
| Horse | 26.80 | 28.16 | 28.05 | 28.32 |
| Mac | 33.73 | 37.55 | 37.83 | 37.57 |
| Allah | 29.28 | 32.94 | 33.14 | 32.61 |
| Old car | 30.60 | 36.00 | 36.45 | 36.29 |
| Lion | 27.16 | 29.49 | 29.62 | 29.72 |
| MRI | 33.92 | 42.28 | 43.26 | 43.06 |
| Einstein | 33.87 | 38.07 | 38.21 | 38.04 |
| Grendizer | 29.59 | 33.07 | 33.24 | 32.78 |

**TABLE II**
**VERTICAL GRID**

|  | Nearest | Linear | Cubic | Kriging |
|---|---|---|---|---|
| Lena | 31.39 | 36.66 | 37.11 | 36.77 |
| Baboon | 24.77 | 27.80 | 27.94 | 27.33 |
| F16 | 30.07 | 37.08 | 37.96 | 36.84 |
| Peppers | 30.22 | 36.09 | 36.77 | 35.97 |
| Mosque | 25.63 | 27.88 | 27.83 | 27.40 |
| Bird | 28.92 | 32.38 | 32.53 | 32.47 |
| Baby | 34.70 | 37.64 | 37.67 | 37.58 |
| Horse | 26.85 | 28.52 | 28.70 | 27.76 |
| Mac | 33.21 | 37.10 | 37.32 | 36.93 |
| Allah | 26.95 | 30.65 | 30.86 | 30.96 |
| Old car | 32.88 | 37.06 | 37.40 | 37.18 |
| Lion | 27.57 | 29.95 | 29.96 | 30.07 |
| MRI | 35.27 | 43.54 | 44.08 | 43.74 |
| Einstein | 34.86 | 40.63 | 40.98 | 40.43 |
| Grendizer | 28.39 | 31.49 | 31.63 | 31.64 |

**TABLE III**
**8-POINTS GRID**

|  | Nearest | Linear | Cubic | Kriging |
|---|---|---|---|---|
| Lena | 26.57 | 27.14 | 28.18 | 28.37 |
| Baboon | 18.85 | 19.26 | 19.48 | 19.93 |
| F16 | 24.23 | 25.08 | 26.25 | 26.83 |
| Peppers | 25.92 | 26.42 | 27.98 | 27.79 |
| Mosque | 19.64 | 20.70 | 20.91 | 21.20 |
| Bird | 23.85 | 24.57 | 25.10 | 25.67 |
| Baby | 30.56 | 31.59 | 32.14 | 30.86 |
| Horse | 23.17 | 23.45 | 23.56 | 23.85 |
| Mac | 28.04 | 29.04 | 29.76 | 30.11 |
| Allah | 22.25 | 23.32 | 23.99 | 23.75 |

| | | | | | | | | | |
|---|---|---|---|---|---|---|---|---|---|
| Old car | 25.67 | 26.55 | 27.77 | 28.23 | | Einstein | 28.40 | 29.15 | 29.71 | 30.22 |
| Lion | 23.17 | 23.69 | 23.88 | 24.24 | | Grendizer | 23.48 | 24.39 | 24.95 | 23.72 |
| MRI | 27.71 | 30.14 | 31.99 | 32.54 | | | | | | |

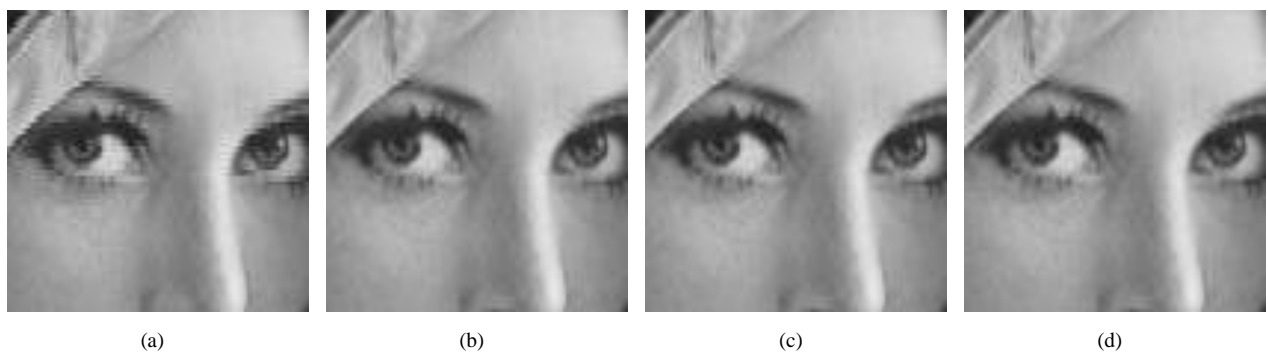

(a)  (b)  (c)  (d)

**Fig. VI. Horizontal grid: (a) Nearest (b) Linear (c) Cubic (d) Kriging**

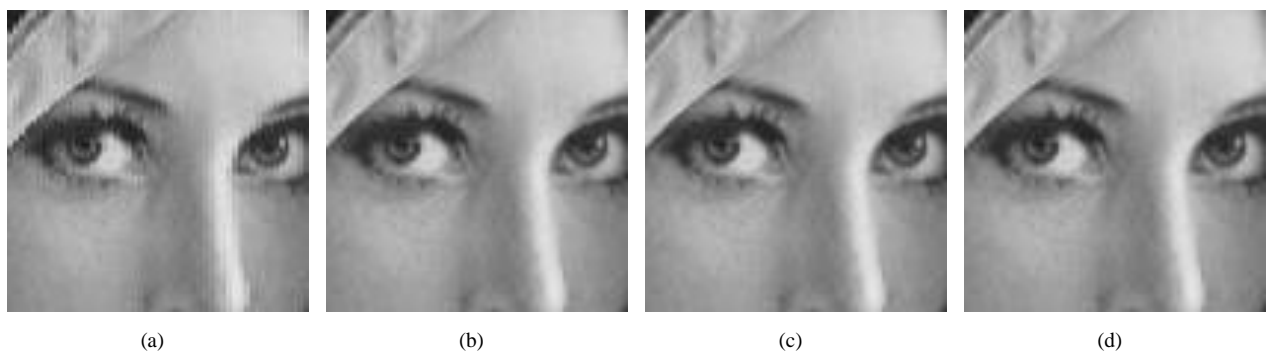

(a)  (b)  (c)  (d)

**Fig. VII. Vertical grid: (a) Nearest (b) Linear (c) Cubic (d) Kriging**

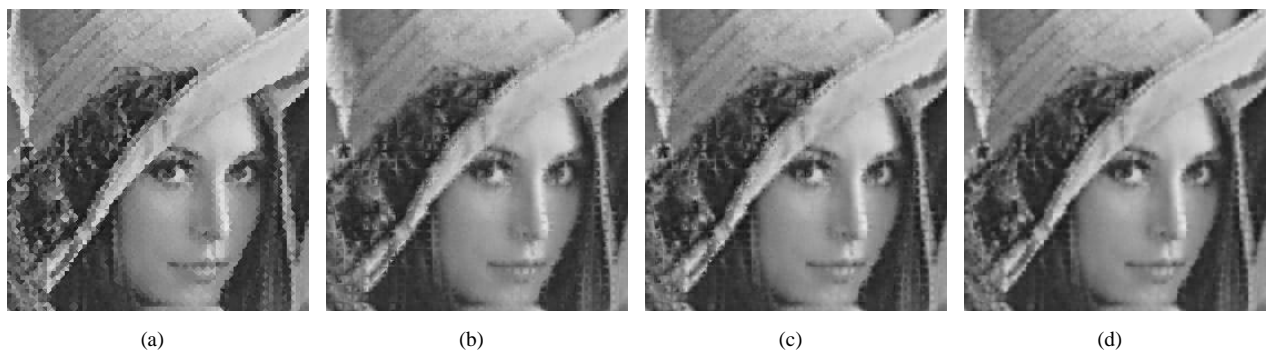

(a)  (b)  (c)  (d)

**Fig. VIII. 8-points grid: (a) Nearest (b) Linear (c) Cubic (d) Kriging**

According to figures VI and VII, it can be seen that the differences between Kriging and classical interpolation, especially cubic spline, are not even noticed by human eye. Whereas in figure VIII it is clearly noticeable that Kriging performs better results than cubic spline and the other traditional methods when used with 8-points gridding method which is the hardest. The ability of Kriging to predict the unknown points from its surrounding neighbors works perfect because it gives weights for each surrounding points in the 8×8 block. The results show that the visual interpolation results can be greatly improved when Kriging interpolation method is adopted.

## V. CONCLUSIONS

In this paper, we propose an efficient algorithm for image interpolation based on Kriging method. Despite kriging being more computationally expensive, it has been shown that it gives better interpolation results when interpolating digital images that have low number of the observed points. Traditional interpolation methods work well when they are prepared with almost high number of observed points. The graphical examples demonstrated that among nearest-neighbor, linear, cubic spline and Kriging methods, the last one give the best performance when used with very little number of the observed original points. Kriging method has competitive efficiency compared with other traditional interpolation methods, and it can be easily extended to any difficult gridding method.

As a future work, the implementation of Kriging method for image interpolation may be applied in image inpainting and resampling.

## BIOGRAPHIES

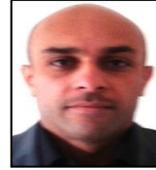

**Firas A. Jassim** was born in 1974, Baghdad, Iraq. He received the BS and MS degrees in mathematics and computer applications from Al-Nahrain University, Baghdad, Iraq in 1997 and 1999, respectively. He received the PhD degree in computer information systems from the University of Banking and Financial Sciences, Amman, Jordan in 2012. His research interests are Image processing, image compression, image enhancement, image interpolation and simulation.

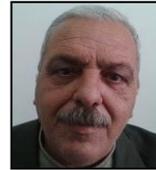

**Fawzi H. Altaany** was born in 1960, Irbid, Jordan. He received the BS degree in public administration from Al-Yermouk University, Irbid, Jordan 1990, and Higher Diploma in health service administration from university of Jordan, 1991 and the MS degree in health administration from Red Sea University, Soudan in 2004 and the PhD degree in management information systems from the University of Banking and Financial Sciences, Amman, Jordan in 2010. His research interests are Image processing, management Information Systems, and customer satisfaction analysis.